\documentclass[letterpaper, 10 pt, conference]{ieeeconf}  

\IEEEoverridecommandlockouts                              

\usepackage{amsmath}
\usepackage{siunitx}
\usepackage{graphicx}
\usepackage{gensymb}
\usepackage{hyperref}
\hypersetup{
   breaklinks=true,   
   colorlinks=false,   
   pdfusetitle=false,  
}
\usepackage{xurl}

\title{\LARGE \bf
A Human-In-The-Loop Simulation Framework for Evaluating Control Strategies in Gait Assistive Robots}

\author{Yifan Wang$^{*1}$, Sherwin Stephen Chan$^{*1,2}$, Mingyuan Lei$^{2}$, Lek Syn Lim$^{2}$ \\Henry Johan$^{2}$, Bingran Zuo$^{2}$, Wei Tech Ang$^{1,2}$
\thanks{*indicates equal contribution}
\thanks{$^{1}$Yifan Wang, Sherwin Stephen Chan and Wei Tech Ang are with School of Mechanical and Aerospace Engineering,
        Nanyang Technological University, 639798 Singapore
        {\tt\small \{ywang114, sherwins001\}@e.ntu.edu.sg}}%
\thanks{$^{2}$Sherwin Stephen Chan, Mingyuan Lei, Lek Syn Lim, Henry Johan, Bingran Zuo and Wei Tech Ang are with the Rehabilitation Research Institute of Singapore, Nanyang Technological University, 308232 Singapore}%
}

\begin{document}

\maketitle
\thispagestyle{empty}
\pagestyle{empty}

\begin{abstract}
As the global population ages, effective rehabilitation and mobility aids will become increasingly critical. Gait assistive robots are promising solutions, but designing adaptable controllers for various impairments poses a significant challenge. This paper presented a Human-In-The-Loop (HITL) simulation framework tailored specifically for gait assistive robots, addressing unique challenges posed by passive support systems. We incorporated a realistic physical human-robot interaction (pHRI) model to enable a quantitative evaluation of robot control strategies, highlighting the performance of a speed-adaptive controller compared to a conventional PID controller in maintaining compliance and reducing gait distortion. We assessed the accuracy of the simulated interactions against that of the real-world data and revealed discrepancies in the adaptation strategies taken by the human and their effect on the human's gait. This work underscored the potential of HITL simulation as a versatile tool for developing and fine-tuning personalized control policies for various users.
\end{abstract}

\section{INTRODUCTION}
Elderly individuals often experience a decline in balance control due to natural aging or geriatric conditions \cite{rosso2013aging}. This impaired balance increases the risk of falls, which are the leading cause of accidental death among the elderly \cite{rubenstein2006falls}. People with balance and mobility impairments require more rehabilitation and assistance in their daily lives. Gait assistive robots \cite{peshkin2005kineassist, mun2015development, wang2011synchronized, marks2019andago,li2023mobile} are promising solutions to address mobility challenges, improve rehabilitation outcomes and ease caregiver burdens. These robots typically feature a mobile base to enable the robot to move with users and a harness system to provide balance support \cite{alias2017efficacy}. Users are encouraged to move independently during gait training, with minimal reliance on robotic assistance, to maximize rehabilitation outcomes. These robots are designed to intervene only when necessary, such as when the user loses balance or is at risk of falling, ensuring both safety and active participation.

Despite the potential benefits, many factors contribute to the slow development and limited adoption of these robots \cite{challenges2022}. Firstly, the heterogeneous nature of gait impairments, stemming from various etiologies and varying severities \cite{moon2016gait}, complicates the design of a universally applicable robot design and controller. Additionally, users exhibit unique adaptive behaviors in response to the device, posing challenges for controller predictability \cite{beckerle2017human}. To ensure assistive robots are truly effective, they must be personalized to meet the specific needs of each user, which demands thorough testing across different condition groups. However, the need for extensive testing introduces ethical and practical challenges, further limiting data-driven optimization and refinement of these devices. 

Human-in-the-loop (HITL) simulation and optimization are increasingly used in assistive robotics to improve controller design and testing \cite{luo2024experiment, gordon2022human, diaz2022human}. HITL simulation addresses these challenges by allowing in-depth physical human-robot interactions (pHRI) analysis in a safe, controlled environment. It simulates human motion and evaluates how interactions with the robot affect user behavior, enabling extensive testing and refinement \cite{folds2015human}.

HITL simulations have been applied in pHRI for tasks such as assisted dressing \cite{clegg2018learning, kapusta2019personalized, clegg2020learning} and grasping \cite{yow2023shared}, where the robot adapts in real-time to a physically separate human. HITL has also been used for exoskeletons—treating the human and robot as a unified system—to develop end-to-end controllers \cite{luo2024experiment}, optimize hip assistance \cite{ding2018human}, and prevent falls \cite{kumar2020learning}. Some sim-to-real frameworks have been proposed for assisted dressing \cite{zhang2022learning} and for model-free exoskeleton controller development \cite{luo2024experiment}, but these sim-to-real transfer focuses primarily on the robot and have minimal evaluation on how the human's movement and behavior has been affected by the robot.

However, current pHRI research focuses on interaction modalities that are either physically separate, as seen in robot arm collaborations, or tightly coupled, as in exoskeleton-based rehabilitation devices. Neither fully addresses the unique requirements of gait assistive robots, which provide a more passive form of support. These systems must balance offering dynamic assistance and allowing the user to maintain as much independence in movement as possible. They only intervene when necessary, such as during balance loss or fatigue, to ensure safety and optimize rehabilitation outcomes \cite{li2023mobile}. For gait assistive robots, achieving high compliance and high transparency is critical to ensuring that the device responds smoothly to the user’s movements while minimizing interference. Yet, these passive systems are not without limitations. Inherent imperfections in their design can introduce unintended effects on the user's gait. For instance, harness friction and lag in the mobile base may cause undesirable push-pull forces, leading to disruptions in natural movement patterns \cite{li2023mobile, mun2014design}. These inconsistencies can hinder rehabilitation by preventing users from engaging in a fluid and natural gait cycle.


As such, this paper presented several key contributions to developing gait assistive robots. Firstly, we presented a HITL simulation framework specifically tailored to gait assistive robots, addressing unique challenges posed by passive support systems. We incorporated a realistic physical human-robot interaction (pHRI) model using a six-DoF mass-spring-damper mechanism to simulate and investigate the human-robot interactions and their effects on human behaviors. Our results highlighted an interesting finding that different adaptation strategies of the human to the robot can yield variations in gait behavior when interacting with the robot. Moreover, the comparison of a speed-adaptive controller and a PID controller through the evaluation of the human-robot interactions underscored the potential of our HITL as a tool for developing customized control strategies tailored to individual users.
\section{HUMAN-IN-THE-LOOP SIMULATION FRAMEWORK}
\begin{figure*}[thpb]
\centering
\includegraphics[scale=0.55]{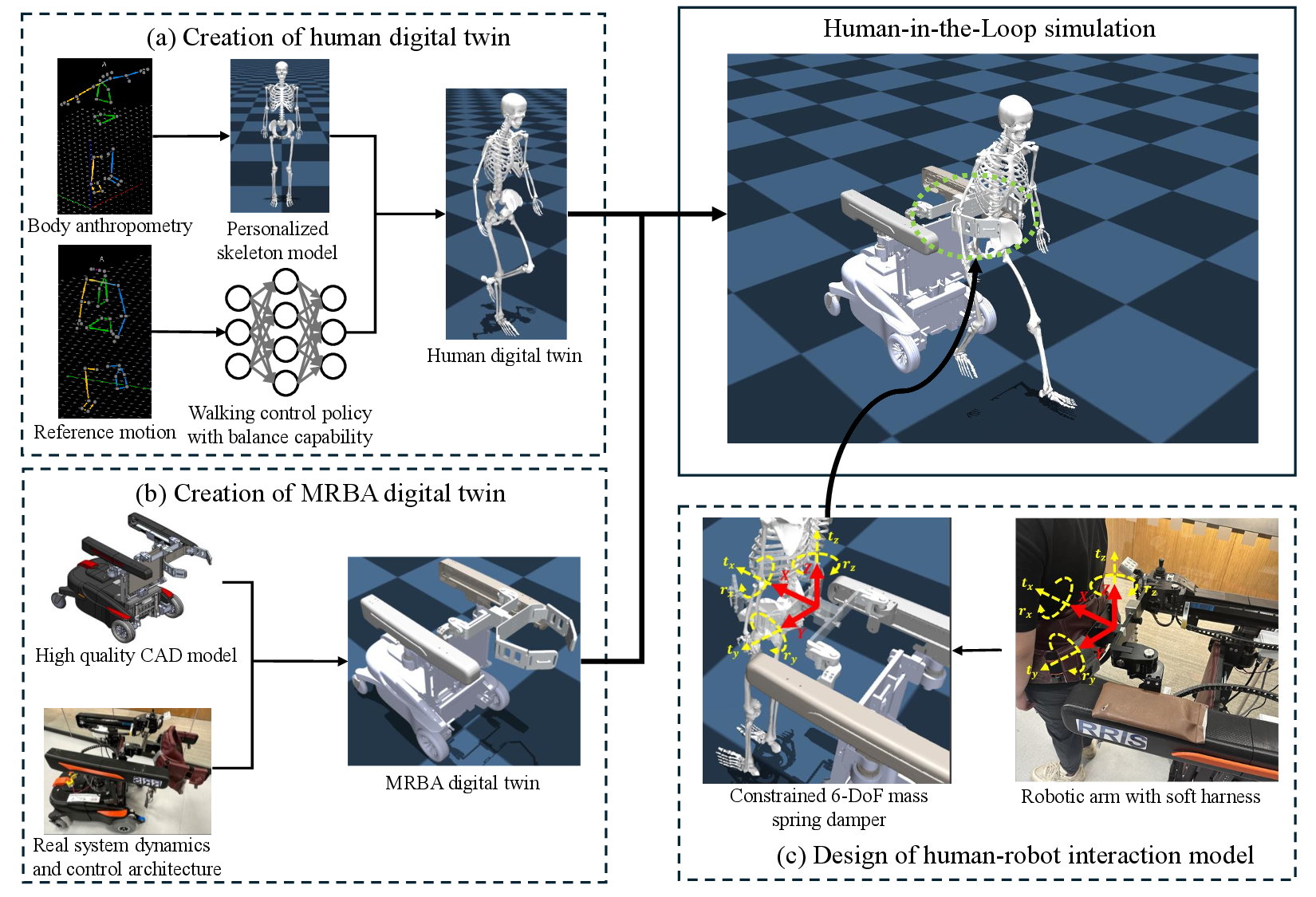}
\caption{Overview of the Human-in-the-Loop (HITL) simulation framework for gait assistive robots. (a) Human digital twin: The process includes using body anthropometry to generate a personalized skeleton model. This model has a walking control policy with varied balance abilities based on the reference motion to create a human digital twin. (b) MRBA digital twin: A CAD model, combined with real system dynamics and control architecture, is used to develop the MRBA digital twin. (c) Physical Human-Robot Interaction (pHRI) model: The human digital twin is constrained with a six-DoF mass-spring-damper mechanism representing the robotic arm with a soft harness.}
\label{fig:HITL_framework}
\end{figure*}
Our HITL simulation framework, based on the MuJoCo physics engine \cite{todorov2012mujoco}, includes three key components: a human digital twin, a robot digital twin, and a physical human-robot interaction model, as shown in Figure \ref{fig:HITL_framework}.
\subsection{Human Digital Twin}
For this work, we defined a full-body skeletal model that contains 27 DoF that is actuated by torque actuators in each joint as seen in Figure \ref{fig:HITL_framework}. Our model has six DoFs un-actuated root joints defined at the pelvis, five ball joints at the (hip, shoulder, and lumbar), and six revolute joints (knee, ankle, elbow) and was built and scaled from a well-documented OpenSim model \cite{rajagopal2016full}. Our 3D personalized human model was created using motion capture data captured via a markerless capture system using the experiment protocol defined by the Rehabilitation Research Institute of Singapore (RRIS) \cite{jatesiktat2024anatomical}. The static data provides information on body anthropometry, while the dynamic data captures baseline gait ability.

We employed a model-free deep reinforcement learning (DRL) approach to develop a human control policy $\pi_h(\textbf{a}_h|\textbf{s}_h)$, where the policy generates an action \textbf{a} based on the state \textbf{s} observed in the environment. We referenced the subject's motions 10m walk test to achieve naturalistic walking in simulation and utilized the network architecture defined by Peng \textit{et al.} \cite{peng2021amp}, leveraging a motion prior and a discriminator to guide the RL agent to learn motions of the reference data. We closely follow their definition of states and discriminator observations. To create varied balance and interactive capabilities for the simulated digital twin, we leveraged our previous work \cite{chan2023creation}, introducing perturbations during training to expose the control policy to diverse disturbances, enhancing both stability and robustness. The simulated gait remains natural throughout this process and closely resembles the subject's reference data.

\subsection{Robot Digital Twin}
The selected gait assistive robot is the Mobile Robotic Balance Assistant (MRBA) \cite{wang2023graceful}, featuring a powered mobile base and a passive robotic arm with three degrees of freedom (DoF). The mobile base follows the user as they walk, while the robotic arm, wrapping around the pelvis, locks in place if balance is lost. MRBA prioritizes a compliant waist interface and a transparent mobile base to minimize disruption to the user’s natural gait while providing necessary support. 

We created the digital twin of the MRBA from a high-quality CAD model, with each component's mass properties and materials carefully defined as seen in Figure \ref{fig:HITL_framework}. System identification of the dynamic parameters, such as joint damping coefficients and friction losses, were estimated and validated in our previous work \cite{wang2023graceful}. These dynamic parameters were then implemented in simulation empirically by manually adding friction losses in each joint so that the simulated model's joints and movement behave similarly to the real robot. The control architecture and parameter settings are identical to the physical robot's, ensuring the digital twin replicates the interactive capabilities of real MRBA: (1) user following (follow-me) and (2) fall detection and intervention. We implemented two follow-me controllers for this work with the first being a typical PID controller, which will be used as a benchmark and the default controller implemented in the deployed robot. However, feedback from post-stroke patient trials revealed the PID controller's inadequacies, with subjects often feeling pushed or pulled due to issues with transparency, harness friction, and input lag. Given these limitations, we aimed to evaluate a second controller developed in \cite{wang2023graceful}. This speed-adaptive controller can adapt to the user's walking speeds in real time and has shown strong compliance with the user's movement in previous non-HITL simulations in MATLAB \cite{MATLAB}. 

\subsection{Physical Human-Robot Interaction Model}
The human pelvis and the robotic arm are connected via a soft harness. While the robotic arm design only allows for three DoF pelvic motions, the natural "slack" between the soft harness and the user’s skin means the user can have some unintended movements. To replicate the pelvis and harness interface between the subject and the robot, we introduced a constrained six DoF mass-spring-damper model to describe the interaction modality, as described by Sun \textit{et al.}, where a linear-spring damper model is appropriate for modeling pHRI interfaces provided individualized parameter tuning is done \cite{sun2023modelling}. As such, a virtual free joint (six DoF) connects the human pelvis and robotic arm as seen in Figure \ref{fig:HITL_framework}. Within the physically plausible joint limits, the relative translational and rotational deviations will increase the interaction forces and torques which are expressed as:
\[ \mathbf{F_i}=\mathbf{k}\cdot\mathbf{q}+\mathbf{d}\cdot\dot{\mathbf{q}} \quad \textit{\text{if}} \quad \mathbf{q}_{min}<\mathbf{q} < \mathbf{q}_{max}\]
where $\mathbf{F}_i=[F_x,F_y,F_z,\tau_x,\tau_y,\tau_z]^{\top}$ is the vector of generated forces and torques. $\mathbf{q}=[q_{tx},q_{ty},q_{tz},q_{rx},q_{ry},q_{rz}]^{\top}$ is the vector of the joint positions of three prismatic joints and three revolute joints.  $\mathbf{k}=[k_{tx},k_{ty},k_{tz},k_{rx},k_{ry},k_{rz}]^{\top}$ and  $\mathbf{d}=[d_{tx},d_{ty},d_{tz},d_{rx},d_{ry},d_{rz}]^{\top}$ represent the joint stiffness and damping coefficients of the respective joints. $\mathbf{q}_{min}=[\underline{q}_{tx},\underline{q}_{ty},\underline{q}_{tz},\underline{q}_{rx},\underline{q}_{ry},\underline{q}_{rz}]^{\top}$ and $\mathbf{q}_{max}=[\overline{q}_{tx},\overline{q}_{ty},\overline{q}_{tz},\overline{q}_{rx},\overline{q}_{ry},\overline{q}_{rz}]^{\top}$ are the lower and upper joint limits. When the joint positions reach the limits, constraint forces are generated to prevent the joint positions from exceeding the defined bounds. 

This representation allows us to tune the allowable movements and flexibility between the human and robot for each degree of freedom. In our implementation, the joint limits were determined based on the experimental observations and are smaller than the natural pelvic deviations observed during walking \cite{neumann2002kinesiology}. After testing and refinement, the joint stiffness and damping coefficient parameters were set empirically.

\section{HUMAN-ROBOT INTERACTION EVALUATION}
Our HITL simulation framework was evaluated by comparing human-robot interaction data from simulation and real-world trials. The main goal is to assess how well the simulation represents real human-robot interactions and the robot’s impact on human behavior, while examining the consistency of control strategy performance across both environments.

\subsection{Experiment Protocol}
A healthy 28-year-old male (95 kg, 1.81 m) participated voluntarily in the experiment. Informed consent was obtained, and the study was approved under IRB-2024-257 by Nanyang Technological University's Institutional Review Board. The subject was instructed to walk naturally at their preferred speed for 10 meters, repeating this task four times. Next, the subject was tightly attached to the robot using the onboard harness. Sufficient time was given to the subject to move and feel comfortable walking with the robot. The subject then walked with the MRBA using the PID and speed-adaptive controllers. For each controller, they completed four 10-meter walks with the robot. Motion data from all trials was recorded using a markerless motion capture system \cite{jatesiktat2024anatomical} with a sampling rate of \SI{50}{\hertz}.

\subsection{Data Processing and Evaluation Metrics}
Four complete gait cycles across four motion capture trials were selected for analysis, resulting in sixteen complete gait cycles for each experiment condition. All kinematic data were processed using a 4\textsuperscript{th}-order Butterworth low-pass filter with a cutoff frequency of \SI{12}{\hertz}. The analysis, both in simulation and real-world conditions, focused on lower limb joint kinematics and spatiotemporal gait parameters. All lower limb joint angles were normalized to the gait cycle.

\subsubsection{Real and Sim Validation}
\begin{figure}[thpb]
\centering
\includegraphics[width=\linewidth, height=0.21\textheight]{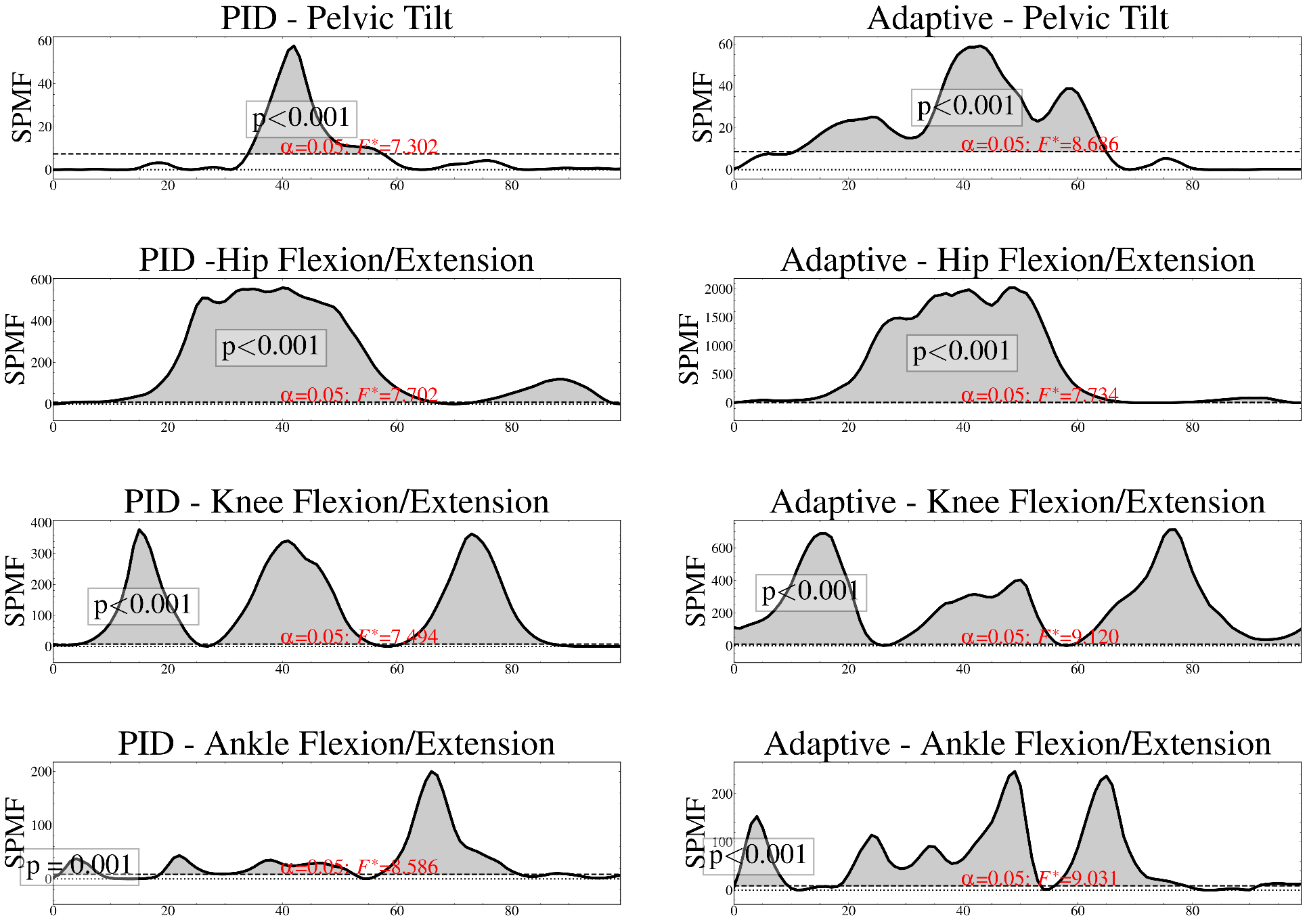}
\caption{Statistical Parametric Mapping (SPM1D) one-dimensional analysis comparing the motion patterns of the subject in simulation and real-world settings.}
\label{fig:spm}
\end{figure}

The validation aimed to evaluate whether the subject exhibits similar motion patterns when interacting with the robot under the same conditions in both simulation and real-world settings. Since human motion is time-series in nature, we applied the Statistical Parametric Mapping (SPM) method, which is widely used in biomechanics research. Specifically, we used a one-way Analysis of Variance (ANOVA) based on random field theory to determine whether there is a significant difference between the two data groups. A threshold is generated, beyond which indicates a significant difference, and the p-value represents the possibility that this difference could result from a smooth random process. In our implementation, the alpha threshold was set at 0.05, and the calculations and plotting were performed using the SPM1D \cite{pataky2012one} Python library, designed for 1-dimensional SPM analysis as seen in Figure \ref{fig:spm}.

\subsubsection{Controller Performance Evaluation}
Traditionally, robot performance evaluations focused on compliance, typically measured by comparing the tracking errors relative to the subject's movements. However, feedback from our clinical trials highlighted the importance of transparency, defined as tolerable interaction forces. Here, we evaluated the PID and speed-adaptive controllers based on both compliance and transparency in real-world and simulation settings, investigating whether the controllers exhibit similar performance across both environments.

(a) Compliance: Since the human performed overground walking tasks, we assessed the tracking errors in two directions -  $e_x$ and $e_y$ representing forward and lateral tracking errors, respectively, as seen in Table \ref{tb:eval_tracking}. The detailed definitions are given in our previous work \cite{wang2023graceful}.

\begin{table}
\caption{Comparison of forward ($e_x$) and lateral ($e_y$) tracking errors}
\label{tb:eval_tracking}
\begin{center}
\begin{tabular}{ccccc}
\hline
Task & \multicolumn{2}{c}{Avg. $|e_x|$ (cm)} & \multicolumn{2}{c}{Avg. $|e_y|$ (cm)}\\
    & Real & Sim & Real & Sim\\
\hline
PID & 4.25 & 7.65 & 3.32 & 7.73\\
Adaptive & 0.37 & 0.7 & 0.72 & 0.53\\
\hline
\end{tabular}
\end{center}
\end{table}

\begin{figure*}[thpb]
\centering
\includegraphics[width=\linewidth]{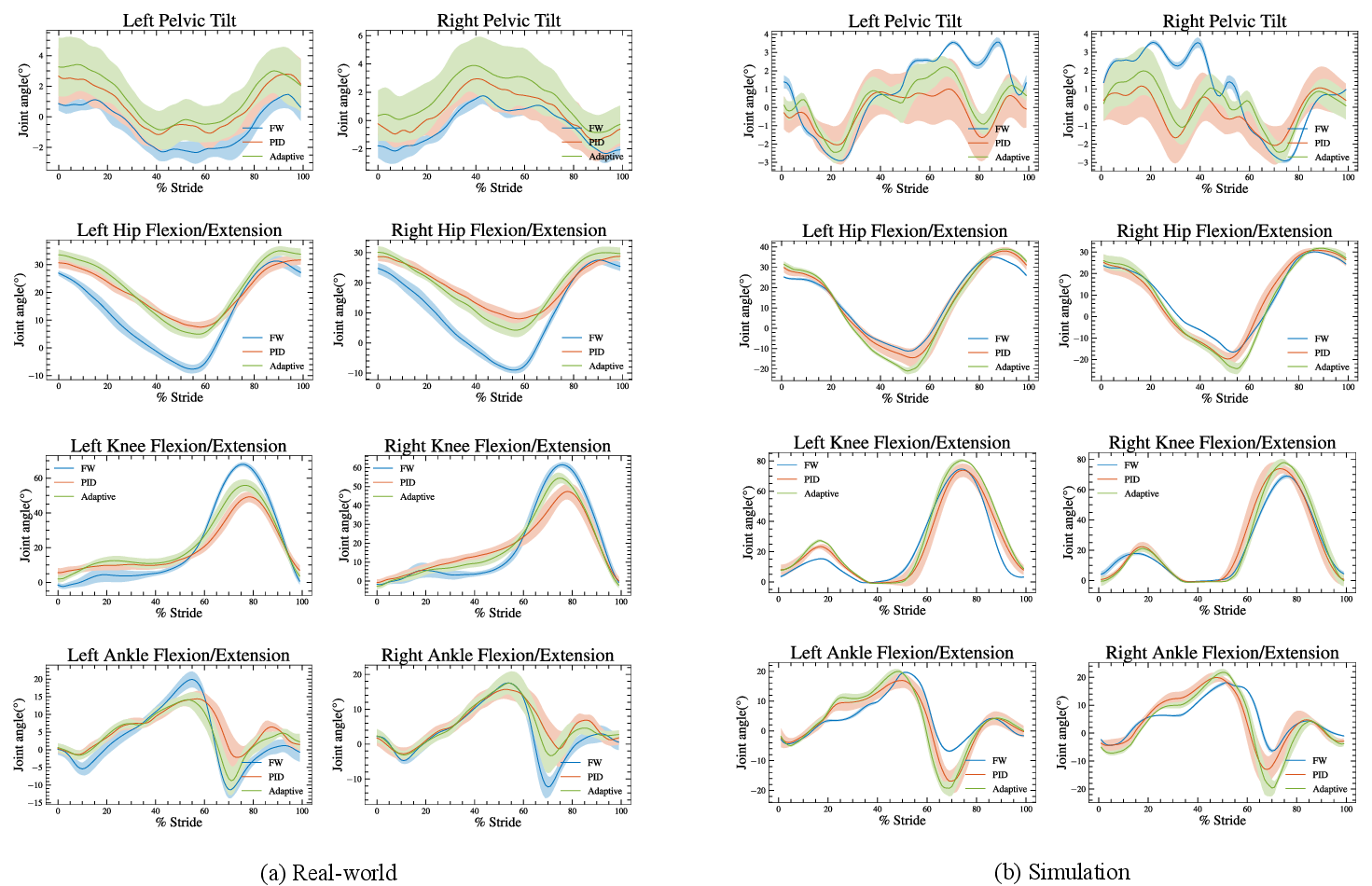}
\caption{Comparison of lower limb joint kinematics for (a) real-world and (b) simulation. The graphs show the joint angles in the sagittal plane for the pelvis, hip, knee, and ankle joints over a complete gait cycle. FW indicates free walking without the robot, PID denotes the use of the PID controller for the follow-me function of MRBA, and Adaptive denotes the speed-adaptive controller for the follow-me function of MRBA. The shaded areas represent the standard deviation of the joint angles across multiple gait cycles.}
\label{fig:joint_kinematics}
\end{figure*}

(b) Transparency: Ideal transparency would result in no interaction forces, which is unrealistic. Additionally, actual interaction forces are difficult to measure in real-world settings. Therefore, we evaluated transparency indirectly by examining whether the subject’s walking dynamics were distorted. Any distortion in walking dynamics would lead to alterations in gait patterns and compromised walking stability. This can be seen by comparing spatiotemporal gait parameters such as stride length and gait speed as seen in Table \ref{tb:eval_gait} and joint angles as seen in Figure \ref{fig:joint_kinematics}. 

\begin{table}
\caption{Comparison of spatiotemporal gait parameters. The values are presented as mean $\pm$ standard deviation}
\label{tb:eval_gait}
\begin{center}
\begin{tabular}{ccccc}
\hline
Task & \multicolumn{2}{c}{Stride length (m)} & \multicolumn{2}{c}{Gait speed (m/s)}\\
    & Real & Sim & Real & Sim\\
\hline
Free walking & 1.29$\pm$0.04 & 1.17$\pm$0.01 & 1.04$\pm$0.02 & 1.12$\pm$0.01\\
PID & 0.90$\pm$0.09 & 1.13$\pm$0.06 & 0.59$\pm$0.06 & 1.28$\pm$0.12\\
Adaptive & 1.14$\pm$0.07 & 1.28$\pm$0.06 & 0.98$\pm$0.02 & 1.47$\pm$0.07\\
\hline
\end{tabular}
\end{center}
\end{table}

Furthermore, we evaluated the transparency of the robot by looking at the walking stability of the subject and compared the standard deviations (SD) of the joint angles across multiple gait cycles as seen in Table \ref{tb:eval_SD} and Figure \ref{fig:joint_kinematics}. A large SD suggests that the subject has to continuously adjust their gait to compensate for the robot's perturbation to restore their walking stability, indicating poorer robot transparency.

\begin{table}
\caption{Comparison of inter-cycle standard deviation of joint angles.}
\label{tb:eval_SD}
\begin{center}
\begin{tabular}{ccccccc}
\hline
Task & \multicolumn{2}{c}{Hip(\degree)} & \multicolumn{2}{c}{Knee(\degree)} & \multicolumn{2}{c}{Ankle(\degree)}\\
    & Real & Sim & Real & Sim & Real & Sim\\
\hline
PID & 1.87 & 2.96 & 3.16 & 4.43& 2.28 &2.67\\
Adaptive & 1.82 & 1.61 & 2.55 & 2.11&2.20 & 1.45\\
\hline
\end{tabular}
\end{center}
\end{table}

\section{RESULTS AND DISCUSSION}
\subsection{Real and Sim Validation}
\subsubsection{Range of Motion Differences}
From Figure \ref{fig:joint_kinematics}(a) from the motion capture data, we observed a reduced range of motion for the hip, knee and ankle in the sagittal plane for both controllers. Comparing this range of motion against the simulated character from Figure \ref{fig:joint_kinematics}(b), we observed no obvious reduction in the range of motion of the hip and knee joint angles in the sagittal plane between the free walking and the two controllers. We posited several reasons for this reduced range of motion at the hip and knee joint angles.

Firstly, despite the robot providing sufficient space in the lower limb region for the subject to maintain their normal range of motion, the real subject may have consciously reduced their step length. This is a pre-emptive move to avoid potential heel collisions with the robot, as the subject cannot perceive where the mobile base following it will be when walking forward. This was not applied in simulation as the simulated character is not environment-aware and the control policy determined that it did not need to account for potential collisions with the robot, thereby maintaining the same range of motion.

Secondly, the waist and thigh harnesses attached also likely contributed to the introduction of resistance and constraints to the natural joint motion. As such, the subject had to reduce their step length, reducing the range of motion for the real subject's hip, knee and ankle joints in the sagittal plane. As for the simulated character, we cannot accurately simulate the various soft straps and harnesses rubbing against the user. Thus, the simulated character would not have any perceived discomfort in the hip region that may cause it to reduce its step length.


\subsubsection{Physical Human-Robot Interaction Differences}
Figure \ref{fig:spm} compared the subject's motions when interacting with the robot between the simulation and real-world settings for both controllers. From the figure, we can observe significant differences, particularly in the hip and knee angles. Since our digital human model was trained without any interaction data with the robot, it independently developed a different adaptation strategy through interaction with the virtual robot. In the real world, we noted that hip angles decreased significantly when walking with the MRBA, but there was no noticeable reduction in the simulation. This reduction was due to the trunk tilting as the user attempted to drag the robot. Biomechanically, hip angles are calculated relative to the trunk, so when the trunk tilts, the hip angles decrease numerically. After accounting for the trunk tilt, no significant differences in hip angles were observed. However, trunk tilting is a natural human adaptation when pulling a heavy object from behind—an adaptation not learned by the digital human model. 

Another difference was observed in the ankle joints after toe-off (60\% of the gait cycle) for both controllers. This difference was caused by excessive dorsiflexion of the ankle joint during toe-off. Biomechanically, this is reasonable because greater ankle dorsiflexion helps generate larger propulsion during walking. This is the adaptation strategy learned by the digital human when walking with MRBA. Since the digital human’s control policy was designed to catch up with the reference motion, it compensated by kicking the ground harder to achieve a greater ground reaction force for propulsion when hindered by the robot from behind. 

The greater ankle dorsiflexion was an unexpected adaptation of the human digital twin's control policy. However, based on our clinical trial observations, this is just one of the many potential adaptations taken by the real subject. As the target users of these gait assistive robots are individuals with gait impairments, impaired users often exhibit weaker lower limb strength, and greater ankle dorsiflexion is unlikely to occur as a compensation strategy. It is more likely that they will lean forward to counteract any dragging force from the robot or reduce their gait speed entirely, adopting a smaller range of motions to cope with the robot's influence. However, not all users adopt a conservative approach. We have also observed users walking faster, feeling more comfortable and confident with the robot’s support \cite{li2023mobile}. This suggests that individual responses to gait assistive robots can vary, with some users adapting in ways that allow them to maintain or even increase their walking speed. 

\subsection{Controller Performance}
Table.~\ref{tb:eval_tracking} compared the tracking errors of the two controllers in the real-world and simulation settings. The speed-adaptive controllers had much smaller tracking errors in two directions, both in the real world and in simulation. Good compliance with the speed-adaptive controller was expected even though this was the first time testing in a HITL simulation scenario. However, both two controllers had better performance in reality than in simulation. The reason is that the human digital twin independently learned a new strategy to compensate for the disturbance from the robot, which made it walk more aggressively in simulation whereas in reality, the subject adapted to the robot more conservatively. This conservative adaptation to the robot led to a more stable coupling dynamically allowing the robot to achieve better compliance.

Table.~\ref{tb:eval_gait} compared stride lengths and gait speeds when walking with the MRBA to free walking for both controllers. In real-world trials, the PID controller led to significant reductions in both stride length and gait speed, which is common in gait assistive robots. This is due to two factors: first, gait assistive robots are typically designed with considerable weight for safety reasons, resulting in high inertia. Second, the PID controller struggles to accommodate changes in walking velocity and gait speed variations among users, leading to slower responses. This delay caused the robot to lag behind the user, forcing them to take shorter steps and reduce their walking speed to adapt. In contrast, the speed-adaptive controller estimated the user's walking speed in real time and adjusted its control strategy accordingly, making the robot more responsive and transparent. As a result, users didn’t need to modify their gait to accommodate the robot. However, both controllers increased the subject’s gait speed in the simulation. This is partly due to the aggressive compensation strategy learned by the digital human, as previously discussed. Additionally, the digital human's body weight was partially supported by the robot, making the lighter virtual body easier to accelerate with the same control policy.

Table.~\ref{tb:eval_SD} compared the standard deviations (SD) of joint angles across different gait cycles for the two controllers. In both the real-world and simulation settings, the speed-adaptive controller resulted in smaller SDs of joint angles compared to the PID controller. This indicates that the speed-adaptive controller causes less distortion to the user's walking dynamics. Additionally, video footage from the simulation revealed that the aggressive compensation strategies of the digital human, combined with the slow response of the robot when using the PID controller, led to a competitive interaction between the two. This significantly compromised the walking stability of the digital human.

\section{CONCLUSIONS AND FUTURE WORK}
In this paper, we proposed a Human-in-the-Loop (HITL) simulation framework specifically tailored for gait assistive robots. Our approach aimed to address the unique challenges of passive support systems by creating digital twins for both the user and the Mobile Robotic Balance Assistant (MRBA). By incorporating realistic physical human-robot interaction (pHRI) models using a six-DoF mass-spring-damper system, we successfully simulated and evaluated the physical human-robot interactions in real-world and virtual settings to provide greater insight into using HITL simulation as a tool for future development of such gait assistive robots.

Our results highlighted key differences between real and simulated human gait behavior when interacting with the robot, particularly in the range of motion and adaptation strategies. The analysis of control strategies showed that the speed-adaptive controller provided better compliance and reduced distortion to the user's natural gait, compared to the conventional PID controller. This finding underscored the potential of our HITL framework in optimizing control strategies for more responsive and transparent interactions between users and gait assistive robots.

Despite this, we noted that some limitations were uncovered. The human digital twin exhibited unexpected adaptations, which pointed to a discrepancy in the learning process of the digital twin. Additionally, the lack of soft straps and soft body simulation inherently limited the fidelity of simulation and dynamic analysis between the physical human-robot interaction.

In our future work, we aim to develop the human digital twin further to learn various compensation strategies, enabling it to adapt to different scenarios beyond just varying balance abilities in the absence of the robot. We also plan to incorporate soft body and soft strap modeling to enhance the realism of the physical interactions, which will improve the simulation's fidelity in representing user movements and constraints. These will provide deeper insights into how different users, especially those with impairments, respond to gait assistive robots and further increase the HITL simulation’s versatility, ultimately enabling the creation and fine-tuning of personalized control policies for each user.




\addtolength{\textheight}{-1cm}   
\section*{ACKNOWLEDGMENT}
This research is supported by A*STAR under the National Robotics Programme (NRP) BAU grant, Assistive Robotics Programme (Award No: M22NBK0074) and Mobile Robotic Balance Assistant (Award No: M23NBK0045). The authors thank the MRBA team for technical assistance and Yan Xiaoyue and Zhang Longbin from Nanyang Technological University for their biomechanics-related discussions.


\bibliographystyle{IEEEtran}
\bibliography{HITL}
\end{document}